\documentclass[conference,a4paper]{IEEEtran}
\IEEEoverridecommandlockouts

\usepackage[hidelinks]{hyperref}
\usepackage[cmex10]{amsmath}
\usepackage{amssymb,amsfonts}
\usepackage{dblfloatfix}

\usepackage[ruled,vlined]{algorithm2e}
\usepackage{graphicx}
\graphicspath{{Figures/PDF/}{Figures/PNG/}}

\usepackage{booktabs}
\usepackage{siunitx}
\usepackage[numbers,compress]{natbib}
\usepackage{texnames}
\usepackage{bm,bbm}
\usepackage{orcidlink}
\usepackage{flushend}

\begin{document}

\title{\uppercase{Hyperspectral in situ remote sensing of water surface nitrate in the Fitzroy River estuary, Queensland, Australia, using deep learning}
\thanks{The authors would like to express their sincere gratitude to CSIRO AquaWatch Mission, CSIRO AI4Missions, CSIRO Data61, CSIRO Environment, CSIRO Earth Analytics Science and Innovation (EASI) platform, and CSIRO AquaWatch in situ sensing team for supporting this work.
}
}

\makeatletter
\newcommand{\linebreakand}{%
  \end{@IEEEauthorhalign}
  \hfill\mbox{}\par
  \mbox{}\hfill\begin{@IEEEauthorhalign}
}
\makeatother

\author{
\IEEEauthorblockN{Yiqing Guo\orcidlink{0000-0002-7179-6267}}
	\IEEEauthorblockA{\textit{CSIRO Data61}\\
		Acton, ACT 2601, Australia\\
		yiqing.guo@csiro.au}
	\and
	\IEEEauthorblockN{Nagur Cherukuru\orcidlink{0000-0002-3617-5852}}
	\IEEEauthorblockA{\textit{CSIRO Environment}\\
		Acton, ACT 2601, Australia\\
		nagur.cherukuru@csiro.au}
	\and
	\IEEEauthorblockN{Eric Lehmann\orcidlink{0000-0001-9145-9551}}
	\IEEEauthorblockA{\textit{CSIRO Data61}\\
		Acton, ACT 2601, Australia\\
		eric.lehmann@csiro.au}
        \and
        \IEEEauthorblockN{S. L. Kesav Unnithan\orcidlink{0000-0002-8758-2554}}
	\IEEEauthorblockA{\textit{CSIRO Environment}\\
		Acton, ACT 2601, Australia\\
		kesav.unnithan@csiro.au}
        \linebreakand
	\IEEEauthorblockN{Gemma Kerrisk\orcidlink{0000-0001-8980-8038}}
	\IEEEauthorblockA{\textit{CSIRO Environment}\\
		Dutton Park, QLD 4102, Australia\\
		gemma.kerrisk@csiro.au}
        \and
	\IEEEauthorblockN{Tim Malthus\orcidlink{0000-0001-7161-8770}}
	\IEEEauthorblockA{\textit{CSIRO Environment}\\
		Dutton Park, QLD 4102, Australia\\
		tim.malthus@csiro.au}
        \and
	\IEEEauthorblockN{Faisal Islam\orcidlink{0000-0003-2610-891X}}
	\IEEEauthorblockA{\textit{CSIRO Environment}\\
		Dutton Park, QLD 4102, Australia\\
		faisal.islam@csiro.au}
}

\maketitle
\begin{abstract}
Nitrate ($\text{NO}_3^-$) is a form of dissolved inorganic nitrogen derived primarily from anthropogenic sources. The recent increase in river-discharged nitrate poses a major risk for coral bleaching in the Great Barrier Reef (GBR) lagoon. Although nitrate is an optically inactive (\emph{i.e.}, colourless) constituent, previous studies have demonstrated there is an indirect, non-causal relationship between water surface nitrate and water-leaving reflectance that is mediated through optically active water quality parameters such as total suspended solids and coloured dissolved organic matter. This work aims to advance our understanding of this relationship with an effort to measure time-series nitrate and simultaneous hyperspectral reflectance at the Fitzroy River estuary, Queensland, Australia. Time-series observations revealed periodic cycles in nitrate loads due to the tidal influence in the estuarine study site. The water surface nitrate loads were predicted from hyperspectral reflectance and water salinity measurements, with hyperspectral reflectance indicating the concentrations of optically active variables and salinity indicating the mixing of river water and seawater proportions. The accuracy assessment of model-predicted nitrate against in-situ measured nitrate values showed that the predicted nitrate values correlated well with the ground-truth data, with an R\textsuperscript{2} score of 0.86, and an RMSE of 0.03 mg/L. This work demonstrates the feasibility of predicting water surface nitrate from hyperspectral reflectance and salinity measurements.
\end{abstract}


\section{Introduction}

The Great Barrier Reef (GBR), extending over 2000 km along the northeastern coastline of Australia, is a renowned coral reef system that has been listed as a World Heritage Area since 1981 by the United Nations Educational, Scientific and Cultural Organisation (UNESCO) as recognition of its ecological significance. Observational evidence suggests that the quality of water in the GBR lagoon has declined as a result of the increase in river discharge of terrestrial nitrogen pollutants from adjacent catchments \citep{macneil2019water, laubenstein2023threats, xiao2023measurement}. As intensified rangeland grazing and agricultural production contribute to excess runoff of eroded surface soil, fertilisers, and pesticides \citep{waterhouse2017}, river discharge of nitrogen to the GBR lagoon is estimated to have increased 2.5--4.5 times from the level prior to European settlement in the 1850s \citep{kroon2012river}. Excess nitrogen is believed to be a major risk factor contributing to the increased susceptibility of corals to bleaching during marine heatwaves. While nitrogen can present in several dissolved inorganic forms, the form of nitrate ($\text{NO}_3^-$, a primary form of dissolved inorganic nitrogen) from anthropogenic sources has been identified as a more significant contributor to increased bleaching prevalence and duration than other forms such as urea or ammonium \cite{donovan2020nitrogen}.

As the largest basin discharging onto the GBR, the Fitzroy catchment has been considered as the region of prioritised interest for monitoring nitrate pollution. Diffuse sources of anthropogenic nitrate have been spotted in the basin, including (1) surface and subsurface erosion caused by rangeland grazing, and (2) fertiliser application in broad-acre cropping especially in sugarcane plantations \citep{kroon2013scientific}. According to the Reef 2050 Water Quality Improvement Plan \citep{queenslandgov2018}, the current level of nitrate in the Fitzroy River needs to be reduced or maintained, in order to meet the whole-of-Reef goal (\emph{i.e.}, over all catchments discharging onto GBR) of reducing anthropogenic end-of-catchment dissolved inorganic nitrogen loads by 60\% by the year 2050.

In response to the increase in river-discharged nitrogen pollution, in-situ instruments have been deployed at the estuary of Fitzroy River by the CSIRO's AquaWatch Australia mission \citep{AquaWatch2024}, enabling real-time measurements of nitrate and associated water quality parameters (\emph{e.g.}, total suspended solids, dissolved organic carbon, and chlorophyll). A hyperspectral sensor, HydraSpectra \citep{HydraSpectra2024}, has been installed at the same location to measure the spectral reflectance of water. The simultaneous measurements of nitrate concentration and water reflectance provide a unique opportunity to study the feasibility of retrieving nitrate from in-situ spectral data in an estuarine environment, with implications for operational mapping of nitrate from spaceborne spectral observations.

In this study, we report the preliminary results on our effort to measure time-series nitrate and simultaneous hyperspectral reflectance at the Fitzroy River estuary, aiming to advance our understanding of their relationship. The rest of this paper is organised as follows. Section \ref{sec:materials_and_methods} introduces the study site, the in-situ instruments, and the modelling procedure for predicting water surface nitrate from hyperspectral reflectance and water salinity measurements. Section \ref{sec:results_and_discussion} presents the modelling results and discusses future directions. Finally, Section
\ref{sec:conclusion} concludes the paper.

\section{Materials and Methods}
\label{sec:materials_and_methods}

\subsection{Study Site}

This study focused on the Fitzroy River estuary, Queensland, Australia, within the Great Barrier Reef (GBR) World Heritage Area. The estuary stretches from the Fitzroy River Barrage near Rockhampton City to the month in Keppel Bay, as shown in Fig.~\ref{fig:estuary_small}. In-situ instruments were set up at a location ($23^\circ 30' 7.44''$S, $150^\circ 48' 3.24''$E) off the coast of Thompson Point (Fig.~\ref{fig:estuary_small}), to measure the water-leaving hyperspectral reflectance and water surface nitrate concentration, as detailed below in Subsections \ref{ssec:hyperspectral_reflectance} and \ref{ssec:nitrate_concentration}. It is worth noting that the study site is a tidal-affected region where nitrate loads are influenced by contributions from both river water and seawater.

\begin{figure}[htb!]
\centering
\includegraphics[width=8.5cm]{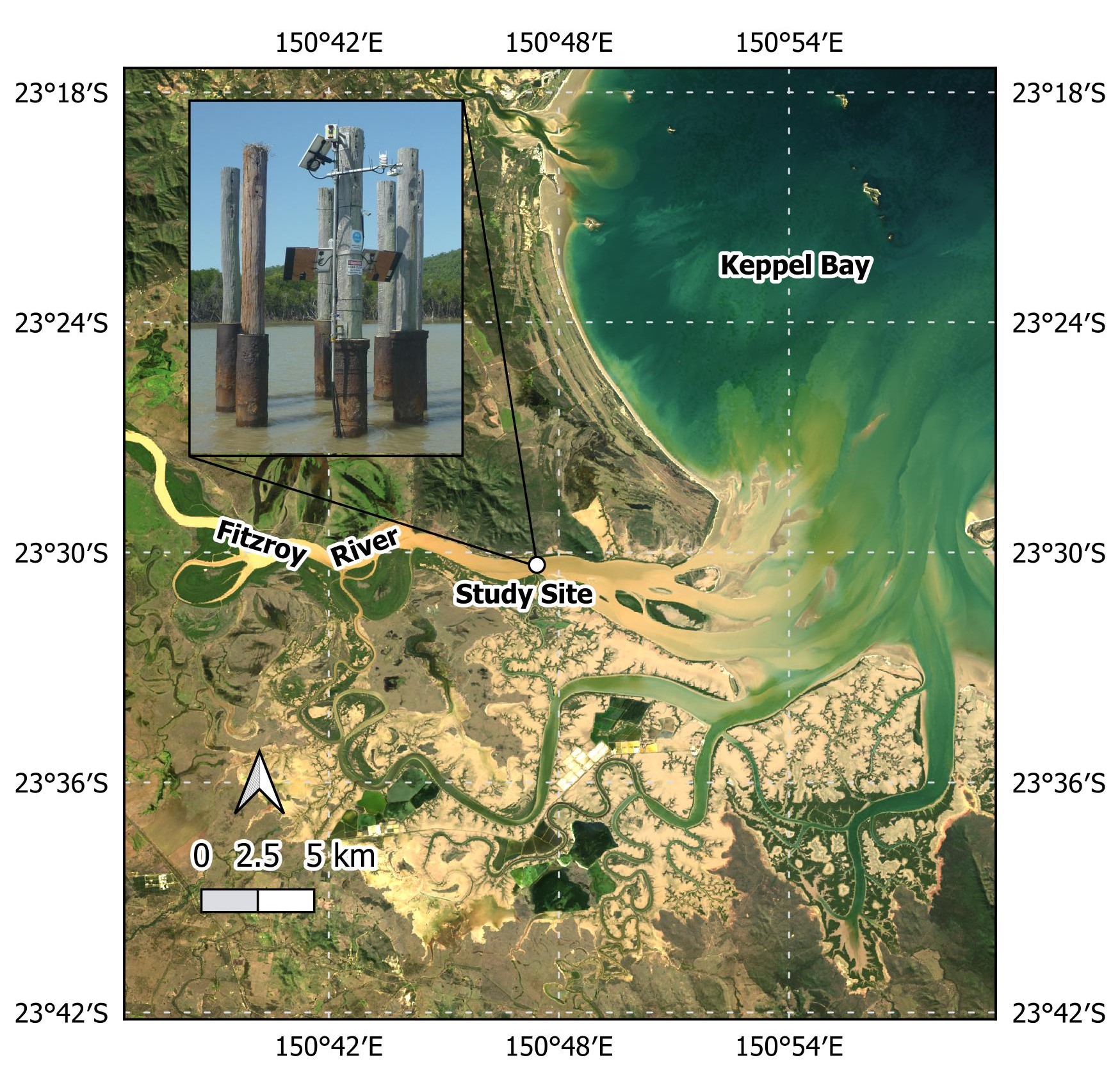}
\caption{Location of the study site. Inset is a photo of the HydraSpectra instrument mounted on the pylon. The background image is displayed using a Sentinel-2 true-color composite, with Bands 4, 3, and 2 mapped to the red, green, and blue color channels, respectively. \label{fig:estuary_small}}
\end{figure}

\subsection{Hyperspectral Reflectance Observations}
\label{ssec:hyperspectral_reflectance}

A HydraSpectra instrument \citep{HydraSpectra2024} was installed at the study site to measure hyperspectral water-leaving reflectance, as shown in Fig.~\ref{fig:estuary_small}. The instrument was calibrated in the laboratory before being mounted onto a pylon at the study site (Fig.~\ref{fig:estuary_small}). It operated autonomously (powered by a battery recharged using solar panels), recording data at a temporal interval of 15 minutes. The Bidirectional Reflectance Factors (BRFs) of water-leaving reflectance were calculated from the spectral measurements of downwelling solar irradiance ($E_d(\lambda)$) and upwelling water-leaving radiance ($L_w(\lambda)$) (Fig.~\ref{fig:hydraspectra}a). The diffuse skylight radiance ($L_{sky}(\lambda)$) was also measured for the correction of skylight reflection (Fig.~\ref{fig:hydraspectra}a). The calculation of BRF followed the method of \cite{mobley1999estimation}. The BRFs recorded at different times of the day were then corrected to a consistent sun-target-view geometry based on a Bidirectional Reflectance Distribution Function (BRDF) simulated by the HydroLight radiative transfer numerical model. Two built-in cameras (Fig.~\ref{fig:hydraspectra}a) operated simultaneously with the spectral sensors to capture true-colour images of the hemispherical sky (Fig.~\ref{fig:hydraspectra}b) and the water horizon (Fig.~\ref{fig:hydraspectra}c), respectively. These true-colour images were leveraged by an analyst to identify and exclude low-fidelity observations, such as those caused by sun glints and bird interference. The retained high-fidelity spectra, which cover the spectral range of 400--750 nm with a spectral resolution of $\sim$5 nm and a spectral sampling interval of 1 nm (Fig.~\ref{fig:hydraspectra}d), were utilised for modelling in our work. Reflectance measurements outside the spectral range of 400--750 nm were removed from our analysis due to the low signal-to-noise ratio.

\begin{figure*}[htb!]
\centering
\includegraphics[width=13.5cm]{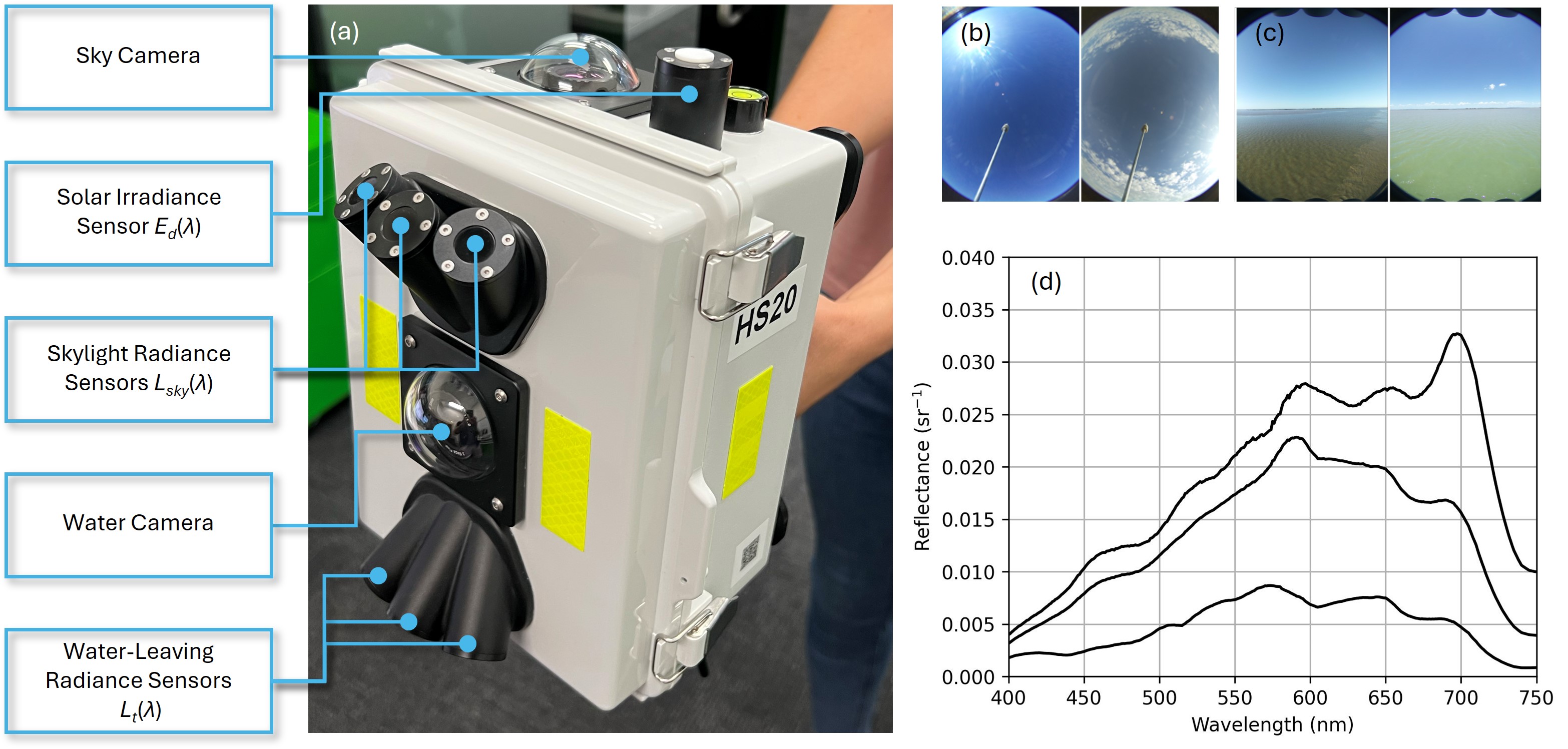}
\caption{(a) The HydraSpectra instrument is equipped with a solar irradiance sensor, three skylight sensors, four water-leaving radiance sensors in the nadir and tilted directions, and two true-colour cameras facing the hemispherical sky and the water horizon. (b) True-colour images recorded by the sky camera. (c) True-colour images recorded by the water camera. (d) Randomly sampled example spectra of water-leaving reflectance recorded by the HydraSpectra instrument. \label{fig:hydraspectra}}
\end{figure*}

\subsection{Nitrate Concentration Measurements}
\label{ssec:nitrate_concentration}

The time-series concentration of nitrate ($\text{NO}_3^-$) was measured using an OPUS UV Spectral Sensor installed approximately two metres below the water surface at the study site. The sensor was powered by a battery, recording a reading of nitrate concentration every 10 minutes with a lab-calibrated accuracy of ±5\% and a precision of 0.005 mg/L. To ensure the quality of the measurements, regular services were conducted for the sensor during the study period. In the event of sensor breakdowns, emergency repairs were carried out soonest possible to minimise data gaps in the time series of measurements. To account for the potential sensor drift that might cause systematic data offsets, a total of 12 in-situ water samples at the study site were collected over the study period, and transported to the CSIRO Hydrochemistry laboratories in Hobart where the nitrate concentrations were analysed. The lab-analysed nitrate samples were then compared with the sensor-recorded nitrate data to cross-calibrate and verify the accuracy of the sensor.

\subsection{Nitrate Concentration Modelling}

A total of 7972 nitrate samples recorded within the time period from October 31\textsuperscript{st}, 2023, to May 6\textsuperscript{th}, 2024 were used for our modelling. Each nitrate observation was matched with a corresponding hyperspectral reflectance measurement recorded within 5 minutes of the nitrate measurement. The paired samples were then randomly divided into a training set (70\%) and a test set (30\%) for model development and accuracy evaluation.

We posit that the relationship between nitrate and spectral reflectance is indirect and non-causal (Fig.~\ref{fig:relation}). Therefore, we estimated the joint distribution between nitrate ($NO_3^-$) and the optically active variables total suspended solids ($TSS$) and coloured dissolved organic matter ($CDOM$), instead of predicting nitrate directly. Both the hyperspectral reflectance ($Rrs$) and salinity ($S$) were used as predictors. The former indicates the concentrations of optically active variables, while the latter indicates the mixing of river water and seawater proportions. Our objective then becomes to estimate the joint probability of $NO_3^-$, $TSS$, and $DOC$ given $Rrs$ and $S$:
\begin{equation}
  P\left(NO_3^-, TSS, DOC \mid Rrs, S\right).  
\end{equation}
We used a conditional Denoising Diffusion Probabilistic Model (cDDPM), denoted as $f\left(\cdot \, ; \, \theta\right)$ with $\theta$ being the model parameters, to derive the maximum likelihood estimation of the above conditional joint distribution. The model parameters $\theta$ are trained by solving the following optimisation problem:
\begin{equation}
\begin{split}
& \hat{\theta} = \arg\min_{\theta} \\
& - \sum_{i=1}^{n} \log f\left( 
{NO_3^-}_{(i)}, {TSS}_{(i)}, {DOC}_{(i)} \mid {Rrs}_{(i)}, S_{(i)} ; \theta  
\right),
\end{split}
\end{equation}
where $\hat{\theta}$ is the estimated model parameters and $n$ is the total number of samples in the training set.

\begin{figure*}[htb!]
\centering
\includegraphics[width=10cm]{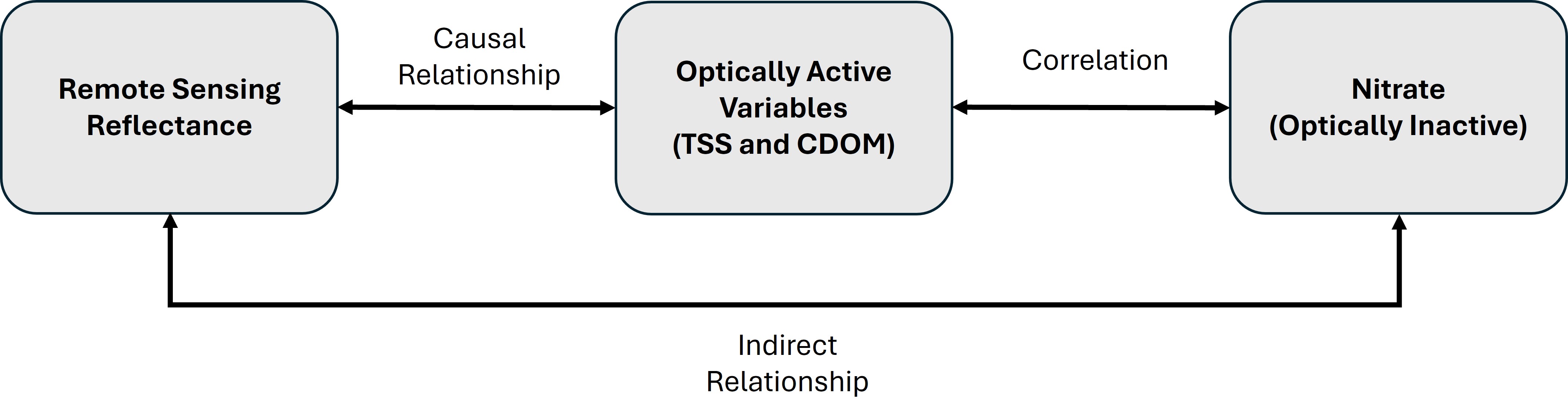}
\caption{The relationship between remote sensing reflectance, optically active water quality variables, and the optically inactive nitrate concentration (TSS: total suspended solids; CDOM: coloured dissolved organic matter).\label{fig:relation}}
\end{figure*}

For accuracy assessment purposes, the coefficient of determination (R\textsuperscript{2}) and root-mean-square error (RMSE) between model-predicted nitrate values and in-situ nitrate measurements were calculated based on the test set.

\section{Results and Discussion}
\label{sec:results_and_discussion}

\subsection{Periodic Cycles of Nitrate}

We observed that the water surface nitrate concentration at the study site was characterised by periodic cycles (Fig.~\ref{fig:daily_cycle}). Figs~\ref{fig:daily_cycle}a and \ref{fig:daily_cycle}b show the time-series nitrate and salinity measurements on April 7\textsuperscript{th} and 15\textsuperscript{th}, 2024, respectively. It was seen that the diurnal pattern of nitrate followed a semidiurnal cycle where the nitrate levels reached their highest and lowest points twice a day. A comparison between Figs~\ref{fig:daily_cycle}a and \ref{fig:daily_cycle}b suggested that the timing of the highest and lowest nitrate readings, as well as the magnitude of daily oscillation in nitrate concentration, varied from date to date. Given that our study site is located within tidal-affect area, these periodic cycles in nitrate loads are believed to be driven by the tidal cycles (Fig.~\ref{fig:daily_cycle}). Nitrate cycles of longer periods, such as those introduced by the alternation of spring tide and neap tide, were also observed in our time-series nitrate measurements. 

\begin{figure}[htb!]
\centering
\includegraphics[width=8.8cm]{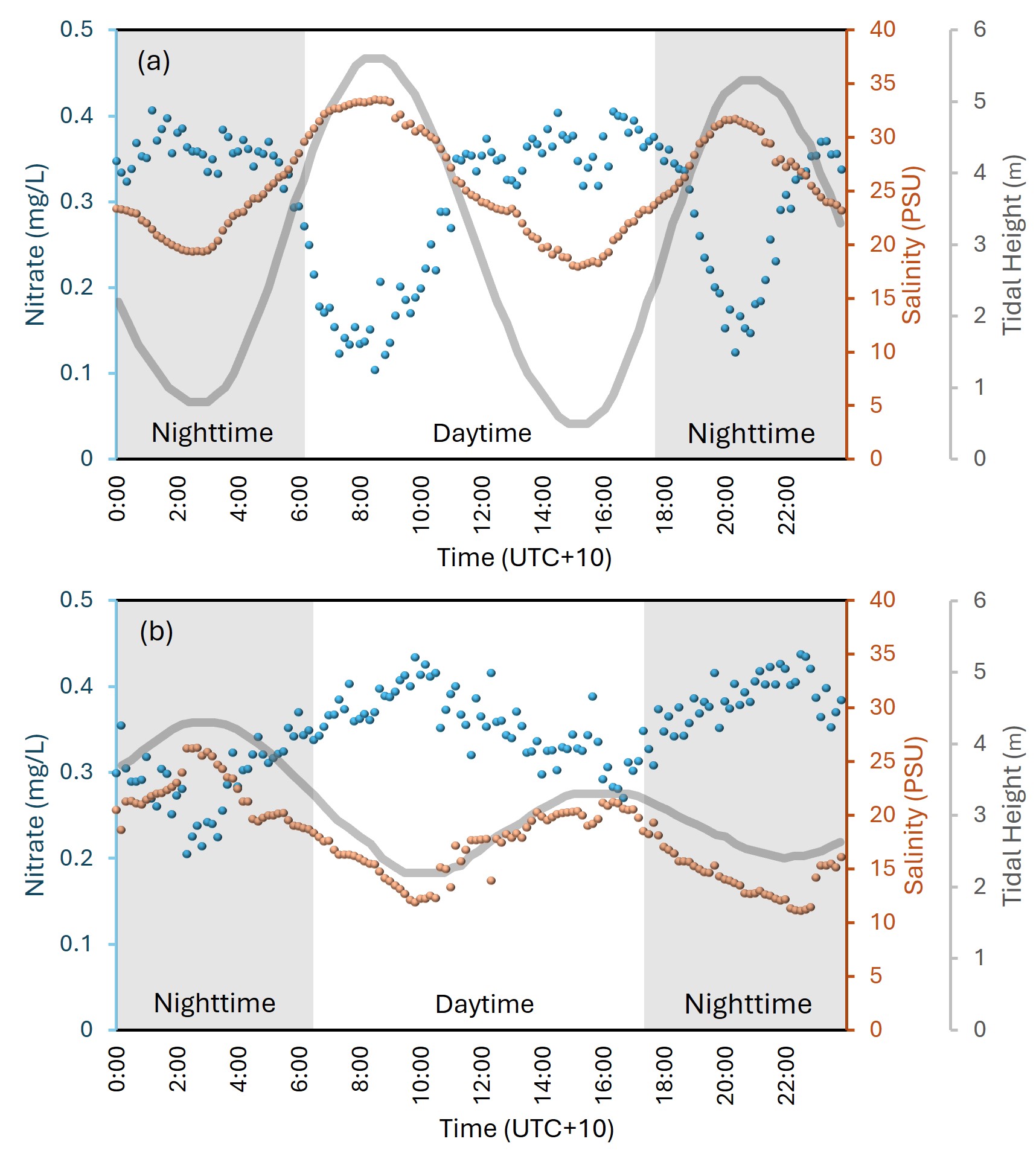}
\caption{Time-series measurements of water surface nitrate concentration (in unit of mg/L) and salinity levels (in unit of PSU) at the study site on (a) April 7\textsuperscript{th}, 2024, and (b) April 15\textsuperscript{th}, 2024. \label{fig:daily_cycle}}
\end{figure}

\subsection{Nitrate Estimation Accuracy}

Figure~\ref{fig:nitrate} shows the accuracy assessment of model-predicted nitrate against in-situ measured nitrate values based on the test dataset, with data points colour-coded based on salinity (PSU). It was observed that the predicted nitrate values correlated well with the ground-truth data,  with an R\textsuperscript{2} score of 0.86, and an RMSE of 0.03 mg/L, suggesting strong model performance in accurately predicting nitrate concentrations at this site. From Fig.~\ref{fig:nitrate} it was seen that the high-salinity waters (mostly seawater) showed lower nitrate concentrations than low-salinity waters (mostly river water), indicating that nitrate loads in river-discharged waters are higher than those in seawater. It is worth noting that the results given in Fig.~\ref{fig:nitrate} are based on our preliminary assessment of the model. Subsequent work could be on future improving the model performance via more extensive validation of the modelling accuracy. 

\begin{figure}[htb!]
\centering
\includegraphics[width=8.8cm]{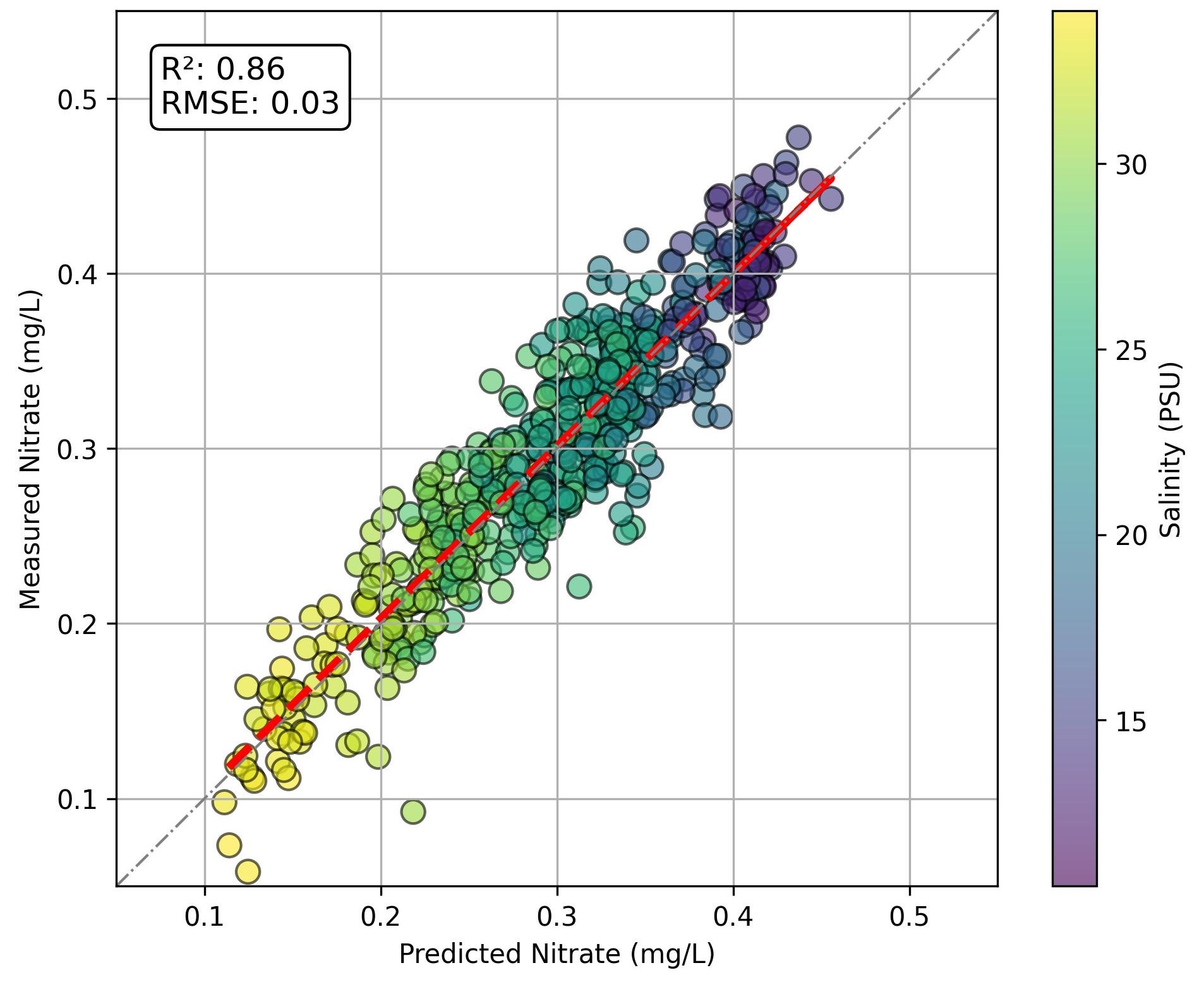}
\caption{Accuracy assessment of model-predicted nitrate against in-situ measured nitrate values. \label{fig:nitrate}}
\end{figure}

\subsection{Future Work}

In this study, we built a model to predict water surface nitrate concentrations from near-surface hyperspectral reflectance and salinity measurements within an estuarine environment, with the predicted nitrate values matching well with ground-truth values. In future work, it is worth further improving the model performance by refining the model structure and its hyperparameters. With the launch of a new generation of hyperspectral satellites, such as PACE \cite{gorman2019nasa}, Tanager \cite{rice2024tanager}, EnMAP \cite{chabrillat2024enmap}, and DESIS \cite{krutz2019instrument, guo2022quantitative, guo2023plant}, the approach proposed in this work has the potential to be up-scaled to satellite hyperspectral observations. Given that our work was focused on the Fitzroy River estuary, future research could also be on extending the proposed method to other regions.

\section{Conclusion}
\label{sec:conclusion}
Monitoring nitrate loads in Fitzroy Estuary helps us understand how much territorial nitrate is discharged onto the Great Barrier Reef lagoon. Given nitrate is optically inactive (\emph{i.e.}, colourless), it is challenging to estimate it from optical measurements of water surface reflectance. In this work, we presented our preliminary efforts on coping with this challenge via modelling the joint distribution of nitrate and the optically active variables TSS and CDOM from hyperspectral reflectance and water salinity. Accuracy assessment of model-predicted nitrate against in-situ measured nitrate values showed that the predicted nitrate values correlated well with the ground-truth data, with an R\textsuperscript{2} score of 0.86, and an RMSE of 0.03 mg/L. This work demonstrates the feasibility of predicting water surface nitrate from hyperspectral reflectance and salinity measurements. Future work would be on further improving the model performance via more extensive validation of the modelling accuracy, expanding the model's application to larger-scale data through up-scaling to spaceborne and airborne hyperspectral observations, and adapting the approach to other geographical regions and environmental conditions.

\small
\bibliographystyle{IEEEtranN}
\bibliography{references}

\end{document}